\documentclass{article}

\usepackage{booktabs}
\usepackage{placeins}
\usepackage{units}
\usepackage{colortbl}
\usepackage{mathtools}
\usepackage{multicol}

\usepackage{color,graphics,epsfig,latexsym,amssymb,amsmath,bm}
\usepackage{graphicx}

\usepackage{listings}
\usepackage{tabularx}
\usepackage[english]{babel}
\usepackage{gensymb}
\usepackage{epstopdf}
\usepackage{pgfplots, pgfplotstable}
\usepackage{tikz-timing}
\usepackage{subcaption}

\usepackage{xspace}

\usepackage{url}
\usepackage{color}
\usepackage[binary-units=true]{siunitx}
\usepackage{listings}
\usepackage{dirtree}
\usepackage{IEEEtrantools}

\usepackage{cprotect}

\usepackage{mwe,tikz}\usepackage[percent]{overpic}
\usepackage{multirow}
\usepackage{subcaption}
\usepackage{enumitem}

\usepackage[rightcaption]{sidecap}

\usepackage{textcomp}
\usepackage{gensymb}
\usepackage{nameref,zref-user, zref-xr}
\usepackage{hyperref}

\hypersetup{
    colorlinks,
    linkcolor={blue!30!black},
    citecolor={blue!30!black},
    urlcolor={blue!30!black}
}

\usepackage{acronym}

\newif\ifMaxQuality

\usepackage[acronym]{glossaries}
\newacronym{DOF}{DoF}{Degree of Freedom}
\newacronym{GPS}{GPS}{Global Positioning System}
\newacronym{UTM}{UTM}{Universal Transverse Mercator}
\newacronym{TI}{TI}{Thermal-Infrared}
\newacronym{LWIR}{LWIR}{Long-Wave Infrared}
\newacronym{NIR}{NIR}{Near-Infrared}
\newacronym{UAV}{UAV}{Unmanned Aerial Vehicle}
\newacronym{CNN}{CNN}{Convolutional Neural Network}
\newacronym{HOG}{HOG}{Histogram of Oriented Gradient}
\newacronym{SVM}{SVM}{Support Vector Machine}
\newacronym{FPN}{FPN}{Feature Pyramid Networks}
\newacronym{SSD}{SSD}{Single Shot Detector}
\newacronym{PDF}{PDF}{Probability Density Function}
\newacronym{SR}{SR}{Systematic Resampling}
\newacronym{PF}{PF}{Particle Filter}
\newacronym{IMU}{IMU}{Inertial Measurement Unit}
\newacronym{IOU}{IoU}{Intersection over Union}
\newacronym{SAR}{SaR}{Search and Rescue}
\newacronym{SWE}{SWE}{Sliding Window Estimator}
\newacronym{JPL}{JPL}{Jet Propulsion Laboratory}
\newacronym{ASL}{ASL}{Autonomous Systems Lab}
\newacronym{ETHZ}{ETHZ}{ETH Zurich}
\newacronym{V4R}{V4R}{Vision for Robotics Lab}
\newacronym{CVG}{CVG}{Computer Vision and Geometry Group}
\newacronym{LiDAR}{LiDAR}{Light Detection And Ranging}
\newacronym{SLAM}{SLAM}{Simultaneous Localization And Mapping}
\newacronym{GNSS}{GNSS}{Global Navigation Satellite System}
\newacronym{DL}{DL}{Deep Learning}
\newacronym{ML}{ML}{Machine Learning}
\newacronym{ICLK}{ICLK}{Inverse Compositional Lucas-Kanade}
\newacronym{DNN}{DNN}{Deep Neural Network}

\newacronym{MDN}{MDN}{Mixture Density Networks}
\newacronym{PVV}{PVV}{Probabilistic Vertex Voting}
\newacronym{AS}{AS}{Active Search}
\newacronym{VAV}{VAV}{Vote-and-verify}
\newacronym{VLAD}{VLAD}{Vector of Locally Aggregated Descriptors}
\newacronym{BOW}{BOW}{Bag-of-Words}
\newacronym{MLE}{MLE}{Maximum Likelihood Estimation}
\newacronym{mm-ISAM}{mm-ISAM}{multi-modal Incremental Smoothing and Mapping}
\newacronym{DEM}{DEM}{Digital Elevation Map}
\newacronym{VI}{VI}{Visual-Inertial}
\newacronym{MI}{MI}{Mutual Information}
\newacronym{SO}{SO}{Ordnance Survey}
\newacronym{ICP}{ICP}{Incremental Closest Point}
\newacronym{kNN}{kNN}{k-Nearest Neighbors}
\newacronym{PnP}{PnP}{Perspective-n-Point}
\newacronym{RANSAC}{RANSAC}{Random Sample Consensus}
\newacronym{GPU}{GPU}{Graphics Processing Unit}

\usepackage{booktabs}

\newcommand{\bft}{\color{blue}\fontseries{b}\selectfont}
\newcommand{\bb}{\color{black}\fontseries{b}\selectfont}

\DeclareMathOperator{\diag}{diag}

\title{Deep UAV Localization \\with Reference View Rendering}
\author{Timo Hinzmann\thanks{All authors are with the Autonomous Systems Lab, ETH Zurich.} \and Roland Siegwart\footnotemark[1]}

\date{}

\begin{document}

%
%
%
\maketitle

\begin{abstract}
This paper presents a framework for the localization of Unmanned Aerial Vehicles (UAVs) in unstructured environments with the help of deep learning.
A real-time rendering engine is introduced that generates optical and depth images given a six Degrees-of-Freedom (DoF) camera pose, camera model, geo-referenced orthoimage, and elevation map.
The rendering engine is embedded into a learning-based six-DoF Inverse Compositional Lucas-Kanade (ICLK) algorithm that is able to robustly align the rendered and real-world image taken by the UAV.
To learn the alignment under environmental changes, the architecture is trained using maps spanning multiple years at high resolution.
The evaluation shows that the deep 6DoF-ICLK algorithm outperforms its non-trainable counterparts by a large margin.
To further support the research in this field, the real-time rendering engine and accompanying datasets are released along with this publication.
\end{abstract}

\section{Introduction}
\gls*{GPS} measurements are commonly fused in aerial navigation systems to guarantee long-term stable navigation.
However, numerous reasons exist for not fully relying on \gls*{GPS} measurements, for instance, flights in naturally \gls*{GPS}-denied areas (e.g., mountainside), intentionally jammed \gls*{GPS} signals from external sources (e.g., by military), unintentionally jammed \gls*{GPS} signals by on-board sources (e.g., USB 3.0 \cite{Lin2014}), or complete \gls*{GPS} receiver failures.
%
%
%
On the other hand, the increasing availability of highly-accurate, high-resolution geo-referenced terrain mosaics, and elevation data allows its tight coupling into a visual-inertial navigation system for \glspl*{UAV}.
The map data is standardized, stored efficiently in the form of images, and can either be acquired from satellite data or generated from geo-referenced images using commercial (e.g., Pix4Dmapper \cite{pix4d}) or open-source software \cite{Hinzmann2017}.
Another advantage of localizing within the map, instead of relying purely on \gls*{GPS}, is that it ensures that the \gls*{UAV} pose and the map are consistent, which makes aerial navigation close to the static terrain safer.
%
However, this raises the challenge of robust image registration under potential environmental and seasonal changes that we want to address in this paper.

\section{Related work}
The related work section is subdivided into the subproblems (a) image retrieval and (b) local image alignment. (a) and (b) are further divided into general localization and \gls*{UAV} specific localization methods.
The remainder of the paper concentrates on the local image alignment, also referred to as map tracking algorithm.
The map tracking algorithm requires a coarse pose initialization, which can be obtained from one of the image retrieval algorithms from Sec.~\ref{sec:global_loc}.
%
%
\subsection{Image Retrieval and Global Localization}\label{sec:global_loc}
\textit{Definition: Given a set of (geo-referenced) database images, retrieve the $N$ best-matching database images for a query image.}

Traditional image retrieval methods can be categorized in \gls*{BOW} and image voting approaches:
FAB-MAP \cite{cummins2008fab} is one of the most famous representant of the \gls*{BOW} approach. 
Proposed by Galvez-Lopez et al., dBoW2~\cite{galvez2012bags} is built up from FAST~\cite{Rosten:Drummond:ECCV2006} and BRIEF~\cite{calonder2010brief} features with focus on quick real-time query.
It is used by the \gls*{VI} state estimator VINS-mono~\cite{qin2018vins} for re-localization.

Many vertex or image voting approaches with different voting strategies exist:
\gls*{PVV} is a scoring scheme proposed by Gehrig et al.~\cite{gehrig2017visual}, that uses BRISK~\cite{leutenegger2011brisk} descriptor projection, approximate \gls*{kNN} search, and scoring based on a binomial distribution.
The strategy significantly outperformed the baseline \gls*{AS}~\cite{sattler2012improving} in an aerial dataset.
The 6-\gls*{DOF} pose is computed with the \gls*{RANSAC} and \gls*{PnP} algorithm based on the identified matches.
Sch\"onberger et al.\ proposed \gls*{VAV}~\cite{schonberger2016vote}, a voting scheme that, in contrast to pure bag-of-words approaches, features spatial verification but aims at being more efficient than standard \gls*{RANSAC} hypothesize-and-verify approaches while maintaining their accuracy.
The image retrieval approach computes a vocabulary and visual words from training images.
Database images are then used for indexing.
For every query, the 6-\gls*{DOF} pose is computed with \gls*{RANSAC}-\gls*{PnP}.
This image retrieval approach is part of the incremental structure-from-motion pipeline colmap~\cite{schoenberger2016sfm}.
Jegou et al. proposed the \gls*{VLAD}~\cite{jegou2010aggregating} descriptor that uses local image features and produces a compact representation for global indexing.
This approach constructs a visual dictionary from the descriptors, computes \gls*{VLAD} descriptors from the visual dictionary, and then creates an index for lookup.
As a first \gls*{DNN}-based global image retrieval method, Arandjelovic et al.\ proposed NetVLAD~\cite{arandjelovic2016netvlad}, a trainable \gls*{CNN} architecture with \gls*{VLAD} layer as main component.
Proposed by Sarlin et al., HF-Net~\cite{sarlin2019coarse} is a hierarchical network based on image retrieval with global descriptors and subsequent 6-\gls*{DOF} pose estimation with local features. 
The approach compresses the Superpoint~\cite{detone2018superpoint} and NetVLAD layers using a knowledge distillation scheme.
Shetty et al. \cite{shetty2019uav} use a \gls*{CNN}-based scene localization network for image retrieval and a camera localization network for pose regression. 
However, this approach is not returning the full 6-\gls*{DOF} pose and results in relatively high position and orientation errors.

Several frameworks exist that explicitly address the challenge of environmental changes between query and database image, with the majority of works designed for automotive or, in general, ground-based applications:
B\"urki et al.~\cite{burki2018map} propose an appearance-based map management system that summarizes multiple missions and improves the query performance while limiting the accumulation of data.
This method is well suited if a location is revisited multiple times with the same sensor setup.
Other approaches make use of semantics to improve invariance to appearance changes: 
Lindsten et al.~\cite{Lindsten2010} estimate the geo-referenced 2D position of a \gls*{UAV} by applying environmental classification to both the terrain mosaic and \gls*{UAV} image stream and using subsequent rotation-invariant template matching for alignment.
Yu et al.~propose VLASE \cite{yu2018vlase}, the aggregation of semantic edges into a feature for subsequent \gls*{VLAD} description for ground vehicle localization.
For ground-aerial localization, X-View~\cite{gawel2018x} creates a multi-view graph of semantic observations but ignores the size and shape of the segments.
Garg et al.~introduce LoST \cite{garg2018lost}, a local semantic tensor combining optical and semantic cues.
Sch\"onberger et al.~\cite{schonberger2018semantic} make use of a 3D semantic descriptor, assuming access to a depth image for every query image.

The majority of methods discussed above treat the queries independently and not as a video stream, with the following exceptions:
Medina et al.~\cite{medina2018video} evaluate different strategies to aggregate image retrieval votes from a video stream to infer the camera's location but do not consider motion between the frames.
X-View \cite{gawel2018x} performs graph matching between the query and database graph using \gls*{MLE}, neglecting the existence of potentially multiple hypotheses.
In contrast, Doherty et al.~\cite{doherty2019multimodal} detect objects of the same semantic class (e.g., cars) and models the multiple data association possibilities with graph-based multi-modal inference~\cite{fourie2016nonparametric}.


\subsection{Local Image Alignment and Map Tracking}
\textit{Definition: Given a query and a target image, compute the transformation that best aligns both images.}

Hand-crafted features and descriptors have been used to match images with small appearance variations:
Koch et al.~\cite{Koch2016} use SIFT~\cite{Lowe:IJCV2004} features with outlier handling to match nearby aerial images. 
A combination of correlation and SIFT-like feature matching is employed in~\cite{Olson2009} to match images recorded of the same terrain but on different days.
Surber et al.~\cite{Surber2017} use weak \gls*{GPS} priors for initialization and direct BRISK matching for visual localization of a \gls*{UAV} in a previously flown mission.
\gls*{MI} methods have shown to be more robust to appearance changes~\cite{Yol2014, Patel2020}, but come with higher computational costs due to the Hessian-based iterative optimization.
%

Higher-level features are employed, for instance, by Patterson et al.\ in~\cite{Patterson2011}.
\gls*{SO} layers with roads and building outlines are used as a reference map.
During the flight, the \gls*{UAV} image stream is converted to the same format, and correlation-based matching is applied to locally align it to the reference map and to estimate the 2D position of the \gls*{UAV}.
Shan et al.~\cite{Shan2017} compute the correlation between \gls*{UAV} image and reference map for global localization. 
After initialization, a particle filter, \gls*{HOG} features, monocular camera, \gls*{IMU}, and barometer are used for local alignment.
Han et al.~\cite{Han2012} extract lines and apply \gls*{ICP} for reference alignment.
Similarly, Son et al.~\cite{Son2009} extract building information from satellite images and match the lines to the ones extracted from the live \gls*{UAV} image stream.

Other methods rely on matching of \glspl*{DEM}: Sim et al.~\cite{Sim2002} estimate the geo-registered \gls*{UAV} position by relative motion estimation from image sequences and absolute position estimation from matching the online computed elevation map to the existing elevation map.
Grelsson et al.~\cite{Grelsson2013} compute the global pose by aligning a geo-referenced 3D model with a local height patch computed from motion stereo.
These approaches assume access to a reliable depth map at run-time and the terrain variation to be discriminative (e.g., urban).
%

The following approaches have been presented that use rendering for reference view generation:
Conte and Doherty use correlation-based template matching fused using a Kalman and Point Mass Filter~\cite{Conte2012}.
Altimeter readings are required for scaling.
Chiu et al.~\cite{Chiu2014} claim map views are rendered from a 3D model but keep the vision module as a black box and focus on the graph-based optimization.
%

Substantial improvements have been achieved with the introduction of \gls*{DL} and \glspl*{CNN}: 
Aznar et al.~\cite{Aznar2016} use a simple \gls*{CNN} for visual alignment of \gls*{UAV} and satellite image but requires a compass and altimeter for operation.
More recent feature and descriptor approaches, such as Superpoint~\cite{detone2018superpoint}, D2-Net~\cite{dusmanu2019d2}, and tracking-based methods like deep \gls*{ICLK} variants~\cite{goforth2018aligning, goforth2019, Lv19cvpr} are able to align two images even under extreme viewpoint and appearance changes.
However, the approach of Goforth et al.~\cite{goforth2018aligning, goforth2019} does not simultaneously compute the alignment and six-\gls*{DOF} geo-referenced pose, and \cite{Lv19cvpr} assumes the availability of depth images for both the target and the reference image, as it is, for instance, the case when using depth cameras.

In this context, this paper introduces a rendering engine that takes geo-referenced orthoimages and elevation maps as input and renders optical and depth images given the six-\gls*{DOF} camera pose and desired camera model.
The rendering engine is combined with a learning-based six-\gls*{DOF}-\gls*{ICLK} algorithm that is able to align images under severe appearance variations and only requires one depth image per image pair.

The remainder of the paper is structured as follows: Sec.~\ref{sec:renderer} introduces the rendering engine that is used for offline training data generation and online reference view generation.
Sec.~\ref{sec:map_tracking} embeds the rendering engine in the learning-based \gls*{ICLK} framework for online six-\gls*{DOF} localization.
The paper concludes with our experiments, a conclusion, and suggestions for future research directions.
%
%
\section{Geo-referenced Renderer}\label{sec:renderer}
This section introduces \verb|aerial_renderer|, a rendering program that uses geo-referenced orthoimages (e.g., optical mosaics of different years) and elevation maps to render a metric depth map and texture images given a query camera pose $\smash{\mathbf{T}^W_C}$, camera intrinsics, and distortion parameters.
The renderer serves two purposes in this paper:
\begin{enumerate}
\item \textbf{Offline training data generation and augmentation:} Parameter inputs are the six-\gls*{DOF} query pose in geo-referenced\footnote{e.g., \gls*{UTM}} coordinates, camera intrinsics, and camera distortion parameters.
The data outputs are the rendered texture images (e.g., from all available optical orthoimages) with the corresponding depth map. 
\item \textbf{Online reference image generation for map tracking (Sec.~\ref{sec:map_tracking}):} 
Given a pose estimate $\smash{\mathbf{T}^W_C}$, a reference image is generated, which is deemed to be close in scale and rotation to the real world image captured by the \gls*{UAV}. 
This has the advantage that the image alignment algorithm \emph{merely} needs to be invariant to seasonal or environmental changes.
Additionally, for the reference image, a dense depth map is rendered essential for six-\gls*{DOF} alignment.
In the online mode, the most recent texture and elevation layers are used for rendering.
\end{enumerate}
%
%

%
\section{Map Tracking}\label{sec:map_tracking}
\begin{figure}[htb]
\includegraphics[width=\linewidth]{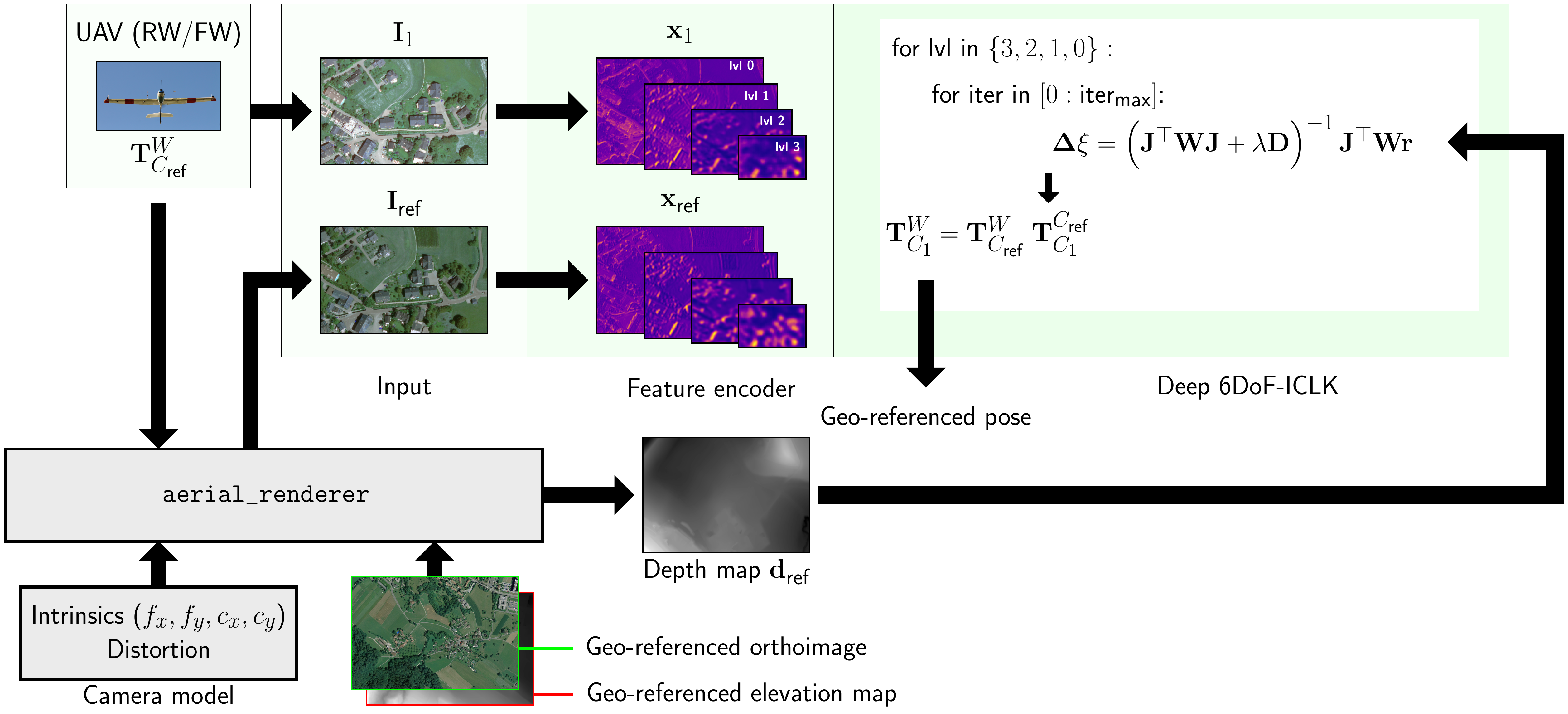}
\caption{Map tracking framework with Deep 6DoF-ICLK.}
\label{fig:map_tracking}
\end{figure}

Fig.~\ref{fig:map_tracking} visualizes our proposed six-\gls*{DOF} \verb|aerial_map_tracker|, combining the geo-referenced rendering engine \verb|aerial_renderer| with a learning-based 6DoF-ICLK algorithm.
The map tracker is agnostic to the \gls*{UAV} platform, i.e., it may be used by rotary-wing (RW) or fixed-wing (FW) \glspl*{UAV}.
In the first step, the current camera pose prior of the \gls*{UAV}, denoted by $\smash{\mathbf{T}^W_{C_\text{ref}}}$, is passed to the rendering engine.
During initialization, this pose may come from a \gls*{GPS}-equipped autopilot, or from one of the coarse global localization algorithms described in Sec.~\ref{sec:global_loc}.
Once initialized, this prior is obtained from the \gls*{ICLK}'s last iteration. 
For the query pose $\smash{\mathbf{T}^W_{C_\text{ref}}}$, the rendering engine returns the optical image $\smash{\mathbf{I}_\text{ref}}$ and depth map $\smash{\mathbf{d}_\text{ref}}$, given the pre-calibrated camera intrinsics and distortion parameters as well as geo-referenced orthoimage and elevation map.
Both, the real optical image taken by the \gls*{UAV}  $\smash{\mathbf{I}_\text{1}}$ and the rendered optical image $\smash{\mathbf{I}_\text{ref}}$ are color normalized and fed as single views to the convolutional feature encoder proposed in \cite{Lv19cvpr}.
The encoder returns four pyramidal layers, ranging from level $0$ ($752\times480$ pixels) to level $3$ ($94\times 60$ pixels).
The \gls*{ICLK} algorithm continues to iterate over all levels, starting with the lowest resolution (level $3$) up to the original resolution image.
For every iteration of the \gls*{ICLK} the six-\gls*{DOF} relative pose is incrementally updated using Levenberg-Marquardt \cite{marquardt:1963, Lv19cvpr}:
\begin{align}
\mathbf{\Delta \xi} = \left(\mathbf{J}^\top\mathbf{W}\mathbf{J}+\lambda \mathbf{D}\right)^{-1}\mathbf{J}^\top\mathbf{W}\mathbf{r}
\end{align}
where the Jacobian $\mathbf{J}$ and residual $\mathbf{r}$ are functions of the depth map $\mathbf{d}_\text{ref}$.
The diagonal matrix $\lambda \mathbf{D} = \lambda \diag{\mathbf{J}^\top\mathbf{W}\mathbf{J}}$ is used for regularization of the Hessian matrix with $\lambda=\smash{1e^{-6}}$.
The weight matrix $\mathbf{W}$ is learned by the convolutional M-Estimator introduced in \cite{Lv19cvpr}.
Once the \gls*{ICLK} reaches a stopping criteria (e.g., max.\ number of iterations) the relative transformation matrix $\mathbf{T}^{C_\text{ref}}_{C_\text{1}}  \in $ SE(3) is computed from $\Delta \xi \in $ se(3) and used to output the desired map-aligned pose $\mathbf{T}^W_{C_\text{1}}$:
\begin{align}
\mathbf{T}^W_{C_\text{1}}=\mathbf{T}^W_{C_\text{ref}}\text{  } \mathbf{T}^{C_\text{ref}}_{C_\text{1}}.
\end{align}
\section{Experiments}
\paragraph{Implementation}
The renderer \verb|aerial_renderer| is implemented in OpenGL (C++) \cite{woo1999opengl}.
Every pixel of the orthoimage is converted into two triangles, i.e., six vertices, inside the vertex shader program.
The \verb|aerial_map_tracker| is implemented in pytorch \cite{NEURIPS2019_9015} and adopted from \cite{Lv19cvpr}.
The \verb|aerial_renderer| is embedded in a ROS \cite{ros} wrapper to interface with \verb|aerial_map_tracker|.

\begin{figure}[htb]
\includegraphics[width=\linewidth]{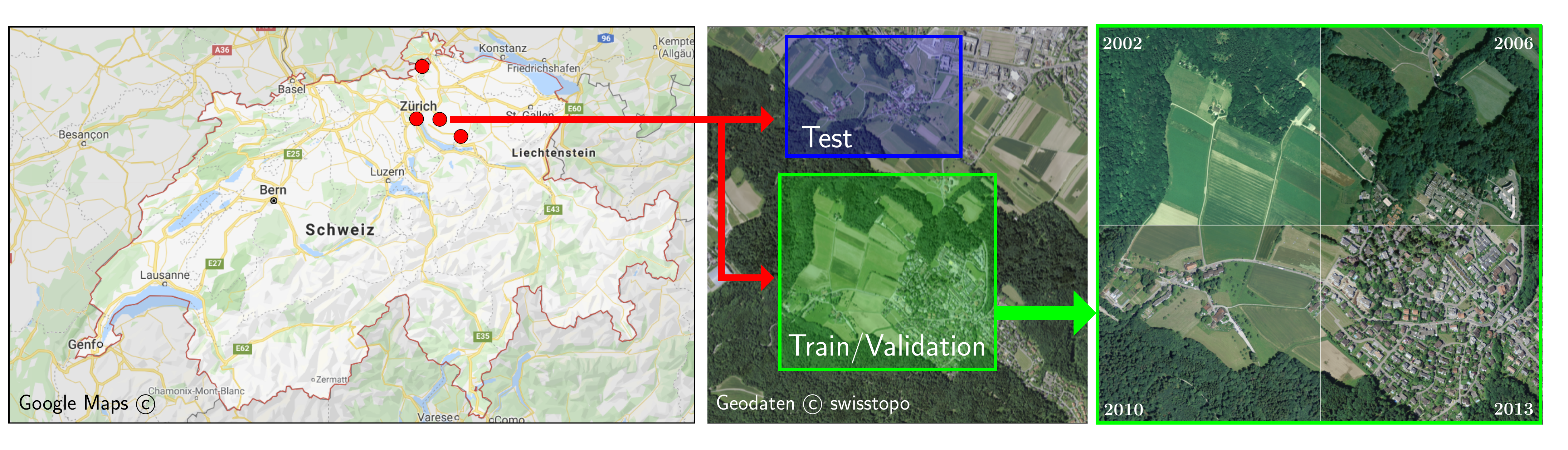}
\caption[]{From left to right: Four locations in Switzerland are selected to extract training, validation and testing data. Training, validation and testing data is extracted for every location from nearby but non-overlapping areas. Each of these areas consists again of orthoimages and elevation maps of all available years.}
\label{fig:training_data}
\end{figure}
\paragraph{Datasets}
Fig.~\ref{fig:training_data} visualizes the four locations in Switzerland from where training, validation, and testing data is extracted.
For each of these four locations, two nearby but non-overlapping areas are selected for training/validation and testing, respectively.
The training and validation set are split $\unit[80]{\%}$ to $\unit[20]{\%}$.
Each of these areas consists again of orthoimages and elevation maps of all available years.
All year-to-year combinations are considered for training (e.g., $2002$ to $2006,2010,2013$).

\paragraph{Results} 
Tab.~\ref{tab:2006_2006} and \ref{tab:2006_2010} present the quantitative results in form of the 3D End-Point-Error (EPE) \cite{Li_2019, Lv19cvpr}, angular, and translational error.
The EPE is used as training loss function for all experiments.
Tab.~\ref{tab:2006_2006} evaluates the alignment for the case that $\smash{\mathbf{I}_\text{1}}$ and $\smash{\mathbf{I}_\text{ref}}$ are rendered from the same year, i.e., same orthoimage.
In contrast, Tab.~\ref{tab:2006_2010} shows the error metrics for two different years (2006, 2010).
\begin{table}[htb]
\resizebox{\textwidth}{!}{\begin{tabular}{l|rrrrr|rrrrr|rrrrr}
\hline
&\multicolumn{5}{c}{\textbf{EPE}}&\multicolumn{5}{c}{\textbf{Ang. Error}}&\multicolumn{5}{c}{\textbf{Transl. Error}} \\
&mean & stdev & median &min &max & mean & stdev & median &min &max & mean & stdev & median &min &max \\ \hline
init & 13.88 & 6.40 & 12.78 & 5.53 & 35.07 & 0.06 & 0.03 & 0.07 & 0.02 & 0.12 & 13.11 & 5.43 & 13.56 & 3.97 & 25.58 \\ \hline
no NN (20) & 7.83 & 8.28 &  7.18 & 0.08 & 35.08 & 0.05 & \bb 0.04 & 0.05 & \bb 0.00 & \bft{0.11} &  8.70 & \bft{6.96} &  8.62 & 0.22 & \bft{26.99} \\
no NN (50)&  3.73 & \bft{8.19} & \bft{0.05} & \bft{0.01} & \bft{34.67} & 0.03 & \bb 0.04 & \bft{0.00} & \bb 0.00 & 0.12 &  4.46 & 7.21 &  \bft{0.18} & \bft{0.01} & 29.21 \\
NN (20) &  \bft{3.19} & 8.55 &  0.58 & 0.10 & 37.82 & \bft{0.02} & \bb 0.04 & 0.01 & \bb 0.00 & 0.18 &  \bft{4.37} & 8.64 &  1.63 & 0.24 & 38.74 \\
\hline
\end{tabular}}
\caption{The error metrics End-Point-Error (EPE), angular and translational error evaluated on the test set for \textbf{the same year} (2006, 2006).}
\label{tab:2006_2006}
\end{table}

First of all, one can see that the deep 6DoF-ICLK algorithm performs on average the best for both cases and in all error metrics, outperforming the variants without trainable parameters independent of the number of iterations.
Secondly, for the same year, the non-trainable 6DoF-ICLK is able to reduce the EPE by a factor of $2.1$ if the number of iterations is increased by a factor of $2.5$, i.e., implying convergence.
\begin{table}[htb]
\resizebox{\textwidth}{!}{\begin{tabular}{l|rrrrr|rrrrr|rrrrr}
\hline
&\multicolumn{5}{c}{\textbf{EPE}}&\multicolumn{5}{c}{\textbf{Ang. Error}}&\multicolumn{5}{c}{\textbf{Transl. Error}} \\
&mean & stdev & median &min &max & mean & stdev & median &min &max & mean & stdev & median &min &max \\ \hline
init & 13.88 &  6.40 & 12.78 & 5.53 & 35.07 & 0.06 & 0.03 & 0.07 & 0.02 & 0.12 & 13.11 &  5.43 & 13.56 & 3.97 & 25.58 \\ \hline
no NN (20) & 14.14 &  \bft{7.87} & 13.90 & 1.41 & \bft{39.48} & 0.08 & \bft{0.04} & 0.08 & 0.01 & \bft{0.16} & 13.85 &  \bft{6.96} & 12.25 & 3.20 & \bft{24.69} \\
no NN (50) & 17.29 & 10.57 & 18.18 & \bft{0.65} & 44.14 & 0.11 & 0.07 & 0.09 & \bft{0.00} & 0.22 & 19.22 & 10.80 & 18.10 & \bft{1.20} & 47.48 \\
NN (20) &  \bft{7.65} &  8.61 &  \bft{5.75} & 1.20 & 40.87 & \bft{0.07} & 0.05 & \bft{0.06} & 0.01 & 0.21 & \bft{10.85} &  7.18 &  \bft{8.96} & 2.88 & 28.70 \\  
\hline
\end{tabular}}
\caption{The error metrics End-Point-Error (EPE), angular and translational error evaluated on the test set for \textbf{different years} (2006, 2010).}
\label{tab:2006_2010}
\end{table}
However, for a different year, the EPE increases by a factor of $1.22$ when increasing the number of iterations from $20$ to $50$, which is a larger error then at initialization, i.e., implying divergence.
The deep 6DoF-ICLK algorithm on the other hand, is able to reduce the EPE well below the initial error in both experiments.

\begin{figure}[htb]
\begin{multicols}{4}
    \includegraphics[width=\linewidth]{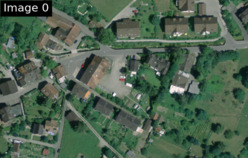}\par
    \includegraphics[width=\linewidth]{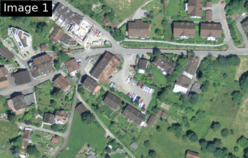}\par
    \includegraphics[width=\linewidth]{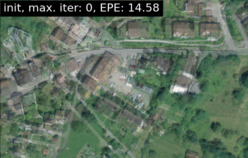}\par 
    \includegraphics[width=\linewidth]{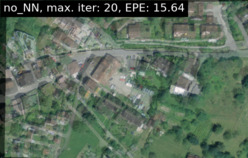}\par 
    \includegraphics[width=\linewidth]{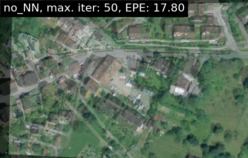}\par 
    \includegraphics[width=\linewidth]{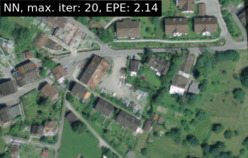}\par   
    \includegraphics[width=\linewidth]{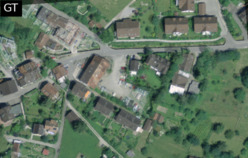}\par
    
    \includegraphics[width=\linewidth]{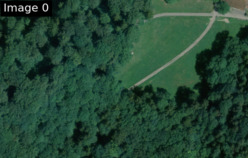}\par
    \includegraphics[width=\linewidth]{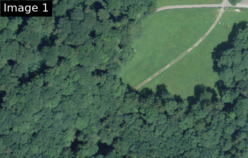}\par
    \includegraphics[width=\linewidth]{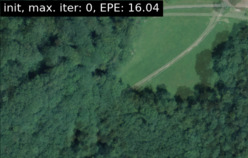}\par 
    \includegraphics[width=\linewidth]{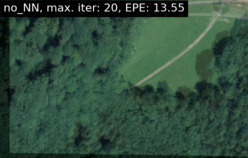}\par 
    \includegraphics[width=\linewidth]{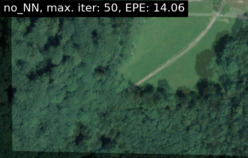}\par 
    \includegraphics[width=\linewidth]{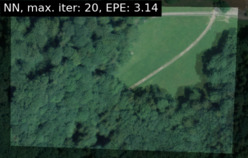}\par  
    \includegraphics[width=\linewidth]{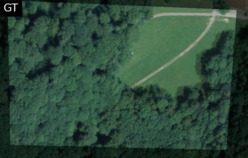}\par
    
    \includegraphics[width=\linewidth]{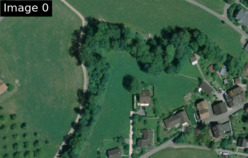}\par
    \includegraphics[width=\linewidth]{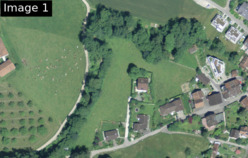}\par
    \includegraphics[width=\linewidth]{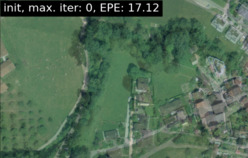}\par 
    \includegraphics[width=\linewidth]{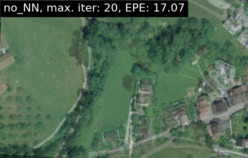}\par 
    \includegraphics[width=\linewidth]{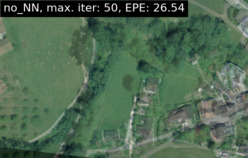}\par 
    \includegraphics[width=\linewidth]{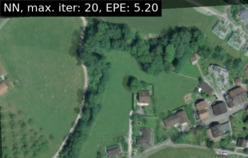}\par  
    \includegraphics[width=\linewidth]{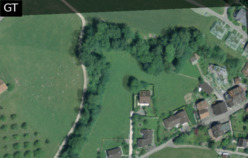}\par    
    
    \includegraphics[width=\linewidth]{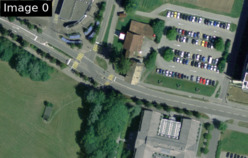}\par
    \includegraphics[width=\linewidth]{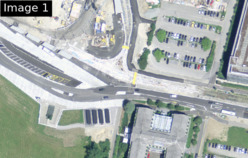}\par
    \includegraphics[width=\linewidth]{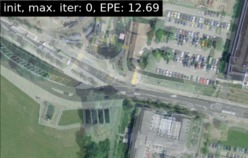}\par 
    \includegraphics[width=\linewidth]{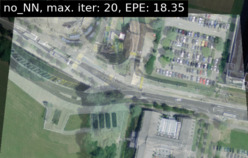}\par 
    \includegraphics[width=\linewidth]{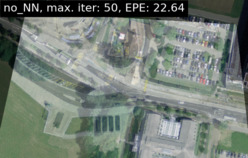}\par 
    \includegraphics[width=\linewidth]{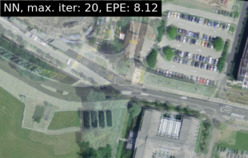}\par  
    \includegraphics[width=\linewidth]{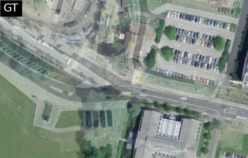}\par    
\end{multicols}
\caption{Qualitative evaluation of four samples from the test set shown in Fig.~\ref{fig:training_data}. The input images image\textunderscore0 and image\textunderscore1 are from the years 2006 and 2010, respectively.}
\label{fig:results_qualitative}
\end{figure}

The qualitative results are visualized in Fig.~\ref{fig:results_qualitative}, presenting four samples taken from the test set (cf.\ Fig.~\ref{fig:training_data}).
The first and second row show the framework input image\textunderscore0 (i.e., $\smash{\mathbf{I}_\text{ref}}$, from year 2006) and image\textunderscore1 (i.e., $\smash{\mathbf{I}_\text{1}}$, from year 2010) respectively.
In the first and, in particular, the last column, one can see significant changes in the infrastructure.
The third row visualizes the overlay of image\textunderscore0 and image\textunderscore1, given the initialized transformation, which is set to identity in all experiments.
The fourth and fifth row show the overlay for the non-trainable 6DoF-ICLK after $20$ and $50$ iterations, respectively, followed by the deep 6DoF-ICLK in the sixth row.
The last row shows the overlay given the ground-truth relative transformation.
All methods are tagged with EPE and the number of iterations.
Overall, the conclusions that can be drawn are consistent with the quantitative analysis described above: In all samples, the EPE for the non-trainable 6DoF-ICLK rises with an increasing number of iterations, and in $3$ out of the $4$ samples it is higher than the initial error.
Instead, the deep 6DoF-ICLK is able to decrease the EPE in all cases drastically.
\setlength{\columnsep}{2pt}

\paragraph{Runtime} 
The \verb|aerial_renderer| runs at over $\unit[2500]{fps}$ on a GeForce RTX 2080 Ti (12GB) for map sizes shown in Fig.~\ref{fig:training_data}.
The runtime of the 6DoF-ICLK algorithm is given in Tab.~\ref{tab:runtime} in $\unit[]{ms}$, tested on the same GeForce RTX 2080 Ti:
It takes the classical 6DoF-ICLK algorithm (\emph{no NN}) approximately $\unit[200]{ms}$ to align one image-pair when stopping after $20$ iterations, which is comparable to the average runtime of the deep 6DoF-ICLK algorithm with the same number of allowed iterations.
Increasing the number of allowed iterations of the classical 6DoF-ICLK algorithm by a factor of $2.5$ increases the runtime by a factor of $2.26$.
\begin{table}[htb]
\scriptsize
\centering
\resizebox{0.65\textwidth}{!}{\begin{tabular}{l|rrrrr}
\hline
&\multicolumn{5}{c}{ \textbf{Runtime} [ms]} \\
&mean & stdev & median &min &max  \\ \hline
no NN (20) &199.30 &  6.07 & 199.72 & 185.97 & 214.77 \\
no NN (50) &450.32 & 12.26 & 446.63 & 421.90 & 476.78 \\
NN (20) &206.85 &  3.67 & 207.51 & 197.99 & 213.31 \\
\hline
\end{tabular}}
\caption{Runtime of the 6DoF-ICLK algorithm in $\unit[]{ms}$.}
\label{tab:runtime}
\end{table}


\section{Conclusion}
This work presents a framework for \gls*{UAV} localization in unstructured environments with the help of \gls*{DL}.
Firstly, this paper introduces \linebreak\verb|aerial_renderer|, a real-time rendering engine that uses geo-referenced orthoimages and elevation maps to render depth and optical images given a six-\gls*{DOF} camera pose $\smash{\mathbf{T}^W_C}$ and camera model.
Secondly, \verb|aerial_map_tracker| is presented, which combines the geo-referenced rendering engine \verb|aerial_renderer| with a learning-based 6DoF-ICLK algorithm.
The evaluation shows that the deep 6DoF-ICLK outperforms its non-trainable counterparts by a large margin.
An interesting alley for future work would be to \emph{hallucinate} the reference image via view synthesis, taking into account important factors like daytime, season, sun position to reduce the environmental or seasonal change and further ease the image alignment process \cite{attribute_hallucination, anokhin2020highresolution}.
Furthermore, we motivate the use of \verb|aerial_renderer| to accelerate the employment of \gls*{DL} for \gls*{UAV} localization in unstructured environments.
For instance, one could create infrared orthomosaics and create infrared-optical or infrared-semantical training data for image alignment.

\section*{Acknowledgments}
The authors thank \emph{Geodaten \textsuperscript{\textcopyright{}} swisstopo} for access to the satellite imagery.
\FloatBarrier
\addcontentsline{toc}{chapter}{Bibliography}

\bibliography{lib.bib}
\bibliographystyle{abbrv}

\end{document}